\documentclass{article}

\usepackage[preprint]{acl}
\usepackage{graphicx}
\usepackage{xcolor}
\usepackage{booktabs}
\usepackage{amsmath}
\usepackage{xurl}

\newcommand{\placeholderbox}[2]{%
  \par\noindent\fbox{%
    \begin{minipage}{\dimexpr\linewidth-2\fboxsep-2\fboxrule\relax}
      \raggedright\textcolor{red}{\textbf{#1:} \url{#2}}%
    \end{minipage}%
  }\par
}
\newcommand{\FIXME}[1]{\placeholderbox{FIXME}{#1}}

\newcommand{\pairedsd}{paired-difference SD}
\newcommand{\marginalsd}{marginal SD}

\providecommand{\result}[1]{\ifcsname nfaresult@#1\endcsname\csname nfaresult@#1\endcsname\else\FIXME{#1}\fi}
\expandafter\newcommand\expandafter{\csname nfaresult@bfcl_gemini_marginal_sd\endcsname}{0.00798146}
\expandafter\newcommand\expandafter{\csname nfaresult@bfcl_gemini_marginal_sd_pp\endcsname}{0.8pp}
\expandafter\newcommand\expandafter{\csname nfaresult@bfcl_8b_marginal_sd\endcsname}{0.00210819}
\expandafter\newcommand\expandafter{\csname nfaresult@bfcl_8b_marginal_sd_pp\endcsname}{0.21pp}
\expandafter\newcommand\expandafter{\csname nfaresult@bfcl_70b_marginal_sd\endcsname}{0.00706233}
\expandafter\newcommand\expandafter{\csname nfaresult@bfcl_70b_marginal_sd_pp\endcsname}{0.71pp}
\expandafter\newcommand\expandafter{\csname nfaresult@bfcl_gemini_paired_noise_floor\endcsname}{0.0112954}
\expandafter\newcommand\expandafter{\csname nfaresult@bfcl_gemini_paired_noise_floor_pp\endcsname}{1.1pp}
\expandafter\newcommand\expandafter{\csname nfaresult@bfcl_8b_paired_noise_floor\endcsname}{0.00282684}
\expandafter\newcommand\expandafter{\csname nfaresult@bfcl_8b_paired_noise_floor_pp\endcsname}{0.28pp}
\expandafter\newcommand\expandafter{\csname nfaresult@bfcl_70b_paired_noise_floor\endcsname}{0.00911825}
\expandafter\newcommand\expandafter{\csname nfaresult@bfcl_70b_paired_noise_floor_pp\endcsname}{0.91pp}
\expandafter\newcommand\expandafter{\csname nfaresult@bfcl_gemini_paired_over_sqrt2_marginal_ratio\endcsname}{1.00}
\expandafter\newcommand\expandafter{\csname nfaresult@bfcl_8b_paired_over_sqrt2_marginal_ratio\endcsname}{0.95}
\expandafter\newcommand\expandafter{\csname nfaresult@bfcl_70b_paired_over_sqrt2_marginal_ratio\endcsname}{0.91}
\expandafter\newcommand\expandafter{\csname nfaresult@bfcl_gemini_mean_pairwise_run_correlation\endcsname}{0.961}
\expandafter\newcommand\expandafter{\csname nfaresult@bfcl_gemini_always_correct_instances\endcsname}{118}
\expandafter\newcommand\expandafter{\csname nfaresult@bfcl_gemini_always_wrong_instances\endcsname}{28}
\expandafter\newcommand\expandafter{\csname nfaresult@bfcl_gemini_sometimes_both_instances\endcsname}{4}
\expandafter\newcommand\expandafter{\csname nfaresult@bfcl_gemini_total_instances\endcsname}{150}
\expandafter\newcommand\expandafter{\csname nfaresult@bfcl_gemini_ever_flip_fraction\endcsname}{2.7\%}
\expandafter\newcommand\expandafter{\csname nfaresult@bfcl_8b_mean_pairwise_run_correlation\endcsname}{0.997}
\expandafter\newcommand\expandafter{\csname nfaresult@bfcl_8b_always_correct_instances\endcsname}{107}
\expandafter\newcommand\expandafter{\csname nfaresult@bfcl_8b_always_wrong_instances\endcsname}{42}
\expandafter\newcommand\expandafter{\csname nfaresult@bfcl_8b_sometimes_both_instances\endcsname}{1}
\expandafter\newcommand\expandafter{\csname nfaresult@bfcl_8b_total_instances\endcsname}{150}
\expandafter\newcommand\expandafter{\csname nfaresult@bfcl_8b_ever_flip_fraction\endcsname}{0.7\%}
\expandafter\newcommand\expandafter{\csname nfaresult@bfcl_70b_mean_pairwise_run_correlation\endcsname}{0.966}
\expandafter\newcommand\expandafter{\csname nfaresult@bfcl_70b_always_correct_instances\endcsname}{121}
\expandafter\newcommand\expandafter{\csname nfaresult@bfcl_70b_always_wrong_instances\endcsname}{26}
\expandafter\newcommand\expandafter{\csname nfaresult@bfcl_70b_sometimes_both_instances\endcsname}{3}
\expandafter\newcommand\expandafter{\csname nfaresult@bfcl_70b_total_instances\endcsname}{150}
\expandafter\newcommand\expandafter{\csname nfaresult@bfcl_70b_ever_flip_fraction\endcsname}{2.0\%}
\expandafter\newcommand\expandafter{\csname nfaresult@study_matched_instances\endcsname}{150}
\expandafter\newcommand\expandafter{\csname nfaresult@study_excluded_instances\endcsname}{0}
\expandafter\newcommand\expandafter{\csname nfaresult@study_unrecoverable_units\endcsname}{0}
\expandafter\newcommand\expandafter{\csname nfaresult@bfcl_gemini_perturbation_accuracy_range\endcsname}{0.797333 to 0.817333}
\expandafter\newcommand\expandafter{\csname nfaresult@bfcl_gemini_perturbation_accuracy_range_pct\endcsname}{80-82\%}
\expandafter\newcommand\expandafter{\csname nfaresult@bfcl_gemini_perturbation_paired_sd_range\endcsname}{0.163212 to 0.208945}
\expandafter\newcommand\expandafter{\csname nfaresult@bfcl_gemini_perturbation_paired_sd_range_pp\endcsname}{16pp to 21pp}
\expandafter\newcommand\expandafter{\csname nfaresult@bfcl_gemini_median_perturbation_paired_sd\endcsname}{0.185462}
\expandafter\newcommand\expandafter{\csname nfaresult@bfcl_gemini_median_perturbation_paired_sd_pp\endcsname}{19pp}
\expandafter\newcommand\expandafter{\csname nfaresult@bfcl_gemini_median_perturbation_over_rerun_paired_sd\endcsname}{16.4192}
\expandafter\newcommand\expandafter{\csname nfaresult@bfcl_gemini_median_perturbation_over_rerun_paired_sd_prose\endcsname}{about 16x}
\expandafter\newcommand\expandafter{\csname nfaresult@bfcl_8b_perturbation_accuracy_range\endcsname}{0.714667 to 0.733333}
\expandafter\newcommand\expandafter{\csname nfaresult@bfcl_8b_perturbation_accuracy_range_pct\endcsname}{71-73\%}
\expandafter\newcommand\expandafter{\csname nfaresult@bfcl_8b_perturbation_paired_sd_range\endcsname}{0.0964432 to 0.186238}
\expandafter\newcommand\expandafter{\csname nfaresult@bfcl_8b_perturbation_paired_sd_range_pp\endcsname}{9.6pp to 19pp}
\expandafter\newcommand\expandafter{\csname nfaresult@bfcl_8b_median_perturbation_paired_sd\endcsname}{0.162915}
\expandafter\newcommand\expandafter{\csname nfaresult@bfcl_8b_median_perturbation_paired_sd_pp\endcsname}{16pp}
\expandafter\newcommand\expandafter{\csname nfaresult@bfcl_8b_median_perturbation_over_rerun_paired_sd\endcsname}{57.6315}
\expandafter\newcommand\expandafter{\csname nfaresult@bfcl_8b_median_perturbation_over_rerun_paired_sd_prose\endcsname}{about 58x}
\expandafter\newcommand\expandafter{\csname nfaresult@bfcl_70b_perturbation_accuracy_range\endcsname}{0.817333 to 0.818667}
\expandafter\newcommand\expandafter{\csname nfaresult@bfcl_70b_perturbation_accuracy_range_pct\endcsname}{82\%}
\expandafter\newcommand\expandafter{\csname nfaresult@bfcl_70b_perturbation_paired_sd_range\endcsname}{0.0516742 to 0.131737}
\expandafter\newcommand\expandafter{\csname nfaresult@bfcl_70b_perturbation_paired_sd_range_pp\endcsname}{5.2pp to 13pp}
\expandafter\newcommand\expandafter{\csname nfaresult@bfcl_70b_median_perturbation_paired_sd\endcsname}{0.0995557}
\expandafter\newcommand\expandafter{\csname nfaresult@bfcl_70b_median_perturbation_paired_sd_pp\endcsname}{10pp}
\expandafter\newcommand\expandafter{\csname nfaresult@bfcl_70b_median_perturbation_over_rerun_paired_sd\endcsname}{10.9183}
\expandafter\newcommand\expandafter{\csname nfaresult@bfcl_70b_median_perturbation_over_rerun_paired_sd_prose\endcsname}{about 11x}
\expandafter\newcommand\expandafter{\csname nfaresult@bfcl_perturbation_order_summary\endcsname}{10.9183 to 57.6315 times larger}
\expandafter\newcommand\expandafter{\csname nfaresult@bfcl_perturbation_order_summary_prose\endcsname}{about 11x to about 58x larger}
\expandafter\newcommand\expandafter{\csname nfaresult@bfcl_perturbation_order_summary_rounded_multiples\endcsname}{11x to 58x larger}
\expandafter\newcommand\expandafter{\csname nfaresult@bfcl_gemini_mean_thoughts_tokens\endcsname}{189}
\expandafter\newcommand\expandafter{\csname nfaresult@bfcl_gemini_wrong_function_failure_count\endcsname}{0}
\expandafter\newcommand\expandafter{\csname nfaresult@bfcl_gemini_wrong_function_failure_rate\endcsname}{0\%}
\expandafter\newcommand\expandafter{\csname nfaresult@bfcl_gemini_wrong_arguments_failure_count\endcsname}{263}
\expandafter\newcommand\expandafter{\csname nfaresult@bfcl_gemini_wrong_arguments_failure_rate\endcsname}{87\%}
\expandafter\newcommand\expandafter{\csname nfaresult@bfcl_gemini_malformed_output_failure_count\endcsname}{2}
\expandafter\newcommand\expandafter{\csname nfaresult@bfcl_gemini_malformed_output_failure_rate\endcsname}{\textless{}1\%}
\expandafter\newcommand\expandafter{\csname nfaresult@bfcl_gemini_other_failure_count\endcsname}{36}
\expandafter\newcommand\expandafter{\csname nfaresult@bfcl_gemini_other_failure_rate\endcsname}{12\%}
\expandafter\newcommand\expandafter{\csname nfaresult@bfcl_8b_wrong_function_failure_count\endcsname}{10}
\expandafter\newcommand\expandafter{\csname nfaresult@bfcl_8b_wrong_function_failure_rate\endcsname}{2\%}
\expandafter\newcommand\expandafter{\csname nfaresult@bfcl_8b_wrong_arguments_failure_count\endcsname}{270}
\expandafter\newcommand\expandafter{\csname nfaresult@bfcl_8b_wrong_arguments_failure_rate\endcsname}{63\%}
\expandafter\newcommand\expandafter{\csname nfaresult@bfcl_8b_malformed_output_failure_count\endcsname}{130}
\expandafter\newcommand\expandafter{\csname nfaresult@bfcl_8b_malformed_output_failure_rate\endcsname}{30\%}
\expandafter\newcommand\expandafter{\csname nfaresult@bfcl_8b_other_failure_count\endcsname}{19}
\expandafter\newcommand\expandafter{\csname nfaresult@bfcl_8b_other_failure_rate\endcsname}{4\%}
\expandafter\newcommand\expandafter{\csname nfaresult@bfcl_70b_wrong_function_failure_count\endcsname}{0}
\expandafter\newcommand\expandafter{\csname nfaresult@bfcl_70b_wrong_function_failure_rate\endcsname}{0\%}
\expandafter\newcommand\expandafter{\csname nfaresult@bfcl_70b_wrong_arguments_failure_count\endcsname}{247}
\expandafter\newcommand\expandafter{\csname nfaresult@bfcl_70b_wrong_arguments_failure_rate\endcsname}{89\%}
\expandafter\newcommand\expandafter{\csname nfaresult@bfcl_70b_malformed_output_failure_count\endcsname}{20}
\expandafter\newcommand\expandafter{\csname nfaresult@bfcl_70b_malformed_output_failure_rate\endcsname}{7\%}
\expandafter\newcommand\expandafter{\csname nfaresult@bfcl_70b_other_failure_count\endcsname}{10}
\expandafter\newcommand\expandafter{\csname nfaresult@bfcl_70b_other_failure_rate\endcsname}{4\%}
\expandafter\newcommand\expandafter{\csname nfaresult@study_endpoint_count\endcsname}{3}
\expandafter\newcommand\expandafter{\csname nfaresult@study_provider_count\endcsname}{2}
\expandafter\newcommand\expandafter{\csname nfaresult@study_suite_count\endcsname}{1}
\expandafter\newcommand\expandafter{\csname nfaresult@study_category_count\endcsname}{2}
\expandafter\newcommand\expandafter{\csname nfaresult@study_perturbation_count\endcsname}{4}
\expandafter\newcommand\expandafter{\csname nfaresult@study_main_temperature\endcsname}{0}
\expandafter\newcommand\expandafter{\csname nfaresult@study_unreported_temperature_arm\endcsname}{0.7}
\expandafter\newcommand\expandafter{\csname nfaresult@study_main_reruns\endcsname}{10}
\expandafter\newcommand\expandafter{\csname nfaresult@study_perturbation_reruns\endcsname}{5}
\expandafter\newcommand\expandafter{\csname nfaresult@sd_curve_n_instance_min\endcsname}{25}
\expandafter\newcommand\expandafter{\csname nfaresult@sd_curve_n_instance_max\endcsname}{150}
\expandafter\newcommand\expandafter{\csname nfaresult@sd_curve_bootstrap_replicates\endcsname}{200}
\expandafter\newcommand\expandafter{\csname nfaresult@bfcl_gemini_sd_curve_full_n_mean_pp\endcsname}{1.3pp}
\expandafter\newcommand\expandafter{\csname nfaresult@bfcl_8b_sd_curve_full_n_mean_pp\endcsname}{0.26pp}
\expandafter\newcommand\expandafter{\csname nfaresult@bfcl_70b_sd_curve_full_n_mean_pp\endcsname}{1.1pp}
\expandafter\newcommand\expandafter{\csname nfaresult@bfcl_gemini_distinct_fingerprints\endcsname}{1}
\expandafter\newcommand\expandafter{\csname nfaresult@bfcl_8b_distinct_fingerprints\endcsname}{11}
\expandafter\newcommand\expandafter{\csname nfaresult@bfcl_70b_distinct_fingerprints\endcsname}{8}

\title{Noise Floor Audit for Agent Benchmarks}

\hypersetup{
  pdftitle={Noise Floor Audit for Agent Benchmarks},
  pdfauthor={Yihang Chen, Pin Qian, Su Wang, Chong Peng, Huan Xu, Xiyang Wu, Yiqi Sun}
}

\author{
  Yihang Chen\\
  \texttt{ychen3726@gatech.edu}
  \And
  Pin Qian\\
  \texttt{pqian@alumni.cmu.edu}
  \AND
  Su Wang\\
  \texttt{suwang@alumni.cmu.edu}
  \And
  Chong Peng\\
  \texttt{chongp@alumni.cmu.edu}
  \AND
  Huan Xu\\
  \texttt{huan.xu71@gmail.com}
  \And
  Xiyang Wu\\
  \texttt{flmbubble@gmail.com}
  \And
  Yiqi Sun\\
  \texttt{ysun697@gatech.edu}
}

\begin{document}

\maketitle

\begin{abstract}
We audit measurement variability for \result{study_endpoint_count} native tool-calling endpoints across \result{study_provider_count} providers on the official BFCL \texttt{multiple} and \texttt{parallel} categories, using matched AST grading.
At temperature \result{study_main_temperature}, reruns are nearly deterministic across Groq endpoints and a thinking-enabled Gemini setting: ever-flip fractions are \result{bfcl_8b_ever_flip_fraction}, \result{bfcl_70b_ever_flip_fraction}, and \result{bfcl_gemini_ever_flip_fraction}, with mean run correlations of \result{bfcl_8b_mean_pairwise_run_correlation}, \result{bfcl_70b_mean_pairwise_run_correlation}, and \result{bfcl_gemini_mean_pairwise_run_correlation}.
Semantics-preserving prompt perturbations create the larger floor on all endpoints, with median perturbation paired SDs \result{bfcl_perturbation_order_summary_rounded_multiples} than rerun paired SDs.
The failure character also shifts: malformed-output failures account for \result{bfcl_8b_malformed_output_failure_rate}, \result{bfcl_70b_malformed_output_failure_rate}, and \result{bfcl_gemini_malformed_output_failure_rate} of task failures, so marginal accuracy hides not only stability but also failure mode.

\end{abstract}

\section{Introduction}

Function-calling benchmarks are increasingly used to compare endpoint quality, reason about deployment readiness, and motivate follow-up research. Their headline scores often appear as single numbers, but a tool-calling run is a compound object: an endpoint emits a structured call, the harness parses provider-specific tool payloads, a grader maps that call to a task outcome, and the reported score aggregates over a finite matched instance set. Reruns at frozen decoding are often treated as a source of measurement noise. This audit measures how large that rerun noise actually is, whether larger variance instead lives in semantics-preserving prompt perturbations, and whether the same marginal score can hide qualitatively different failure modes.

This paper is a noise floor audit rather than a new benchmark. We focus on the official Berkeley Function Calling Leaderboard categories used in the frozen study set, where native tool calling, schema adherence, argument formatting, and AST grading define the measurement pipeline. The completed audit covers \result{study_endpoint_count} endpoints across \result{study_provider_count} providers, including a thinking-enabled Gemini setting with mean thinking-token use of \result{bfcl_gemini_mean_thoughts_tokens} tokens per retained main-arm call. The goal is to make uncertainty visible at the level at which benchmark claims are usually made.

Our contributions are:

\begin{itemize}
  \item We find that reruns at frozen decoding are nearly deterministic across the \result{study_endpoint_count} endpoints: ever-flip fractions are \result{bfcl_8b_ever_flip_fraction}, \result{bfcl_70b_ever_flip_fraction}, and \result{bfcl_gemini_ever_flip_fraction}, with mean run correlations of \result{bfcl_8b_mean_pairwise_run_correlation}, \result{bfcl_70b_mean_pairwise_run_correlation}, and \result{bfcl_gemini_mean_pairwise_run_correlation}.
  \item We show that semantics-preserving prompt perturbations create a much larger floor on all endpoints: median perturbation paired SDs are \result{bfcl_perturbation_order_summary_rounded_multiples} than rerun paired SDs, a pattern visible only under matched paired analysis.
  \item We find that endpoint capability does not consistently predict measurement stability: across the two Groq sizes, rerun stability and perturbation stability rank the endpoints in opposite orders, and the cross-provider, thinking-enabled Gemini point confounds provider and thinking configuration rather than extending a clean scale axis.
  \item We add a failure taxonomy showing that malformed-output failures account for \result{bfcl_8b_malformed_output_failure_rate}, \result{bfcl_70b_malformed_output_failure_rate}, and \result{bfcl_gemini_malformed_output_failure_rate} of task failures. The character of failure shifts from structural failures toward well-formed-but-wrong calls, a distinction hidden by marginal accuracy.
\end{itemize}

\section{Related Work}

\paragraph{Evaluation variance and error bars.}
Large language model evaluations have made broad capability comparisons easier to communicate, but they also amplify the need to report uncertainty around aggregate scores. Dror et al. give NLP-specific guidance for selecting statistical significance tests, while Dodge et al. show that reporting only the final test-set score can hide instability caused by development choices \citep{dror2018hitchhikers,dodge2019show}. Miller gives a direct proposal for adding uncertainty estimates to language-model evaluations, including paired comparisons and evaluation planning \citep{miller2024errorbars}. Our audit adds a narrower empirical target: estimate rerun and prompt-perturbation variability for repeated tool-calling evaluations under a fixed harness.
Recent benchmark-audit work makes related measurement concerns explicit in other settings: configuration choices alone can reverse pairwise alignment-benchmark rankings, and paired-MDE planning can make detectable-effect assumptions explicit before running quantization benchmarks \citep{li2026safetyreproconfigurationconditionalrankinstability,zhuang2026preregisteringdetectableeffectpairedmde}. Standardization work likewise shows that prompt formatting, in-context examples, probability normalization, task formulation, and benchmark signal-to-noise can materially affect reproducible model comparisons \citep{gu2025olmes,heineman2025signal}. We study a different measurement object, but use the same basic premise that benchmark claims should expose the uncertainty induced by the evaluation design.
Cross-domain evaluation audits make related bounded points: observation-window sufficiency in churn prediction can depend on cohort, target, and feature design, while predictive-uncertainty work separates epistemic and aleatoric components rather than collapsing uncertainty into one scalar \citep{han2026earlyearlyenoughdesigndependent,fu2026sphunchypersphericaluncertaintydecomposition}. NFA's components are different: measurement-layer rerun variation and prompt-surface variation in a fixed native tool-calling harness.

\paragraph{Reproducibility.}
Reproducible language-model evaluation depends on endpoint versioning, decoding controls, retry policy, parser behavior, grader implementation, and raw-output retention. Biderman et al. document recurring sources of irreproducibility in language-model evaluation and motivate more explicit harness conventions \citep{biderman2024trenches}. Our audit adopts that reproducibility framing but focuses on a single question: how much score movement remains after the suite, prompt template, grader, and matched instances are held fixed?

\paragraph{Prompt sensitivity and surface-form robustness.}
Prompt robustness work has shown that semantic equivalence does not imply behavioral invariance. Formatting-only changes can move few-shot task accuracy by large amounts, and multi-prompt evaluation studies show that paraphrased instruction templates can change both absolute scores and relative model rankings \citep{sclar2024format,mizrahi2024state}. Other work measures robustness to prompt perturbations that preserve semantic integrity or defines sensitivity and consistency metrics over prompt rephrasings \citep{zhu2024promptrobust,errica2025wrong}. Prompt Privilege studies a related user-facing problem: expert-style prompts can receive better answers than lower-literacy formulations of the same intent \citep{jin2026same}.
This noise floor audit (NFA) studies a different measurement object: matched native tool-calling benchmark outcomes under declared semantics-preserving prompt perturbations, where the goal is to estimate benchmark measurement sensitivity rather than optimize prompting, measure adversarial robustness, or quantify user-access inequity.

\paragraph{Failure-mode analysis.}
Several evaluation lines argue that scalar accuracy can obscure the character of model failures.
AgentBoard reports fine-grained progress and multifaceted agent analyses rather than only final success rates \citep{ma2024agentboard}.
Adjacent audits make a similar point outside tool calling: trace-level bias-acknowledgment diagnostics separate responses with the same final-answer score, regime-stratified forecasting evaluation exposes transition failures hidden by aggregate metrics, and generated-video evaluation shows that binary fake/real labels can be easier than fine-grained trace identification, grounding, and explanation \citep{sun2026beyond,wang2026timeseriesfoundationmodel,fu2025learninghumanperceivedfakeness}.
For tool-use systems, T-Eval decomposes tool utilization into subprocesses such as instruction following, planning, retrieval, understanding, and review, while ToolScan characterizes tool-use error patterns including incorrect function names, argument names, argument values, argument types, and invalid formats \citep{chen2024teval,kokane2025toolscan}.
Recent work on agent execution provenance makes the same final-answer limitation explicit for longer-horizon agents, arguing that trace and evidence records are needed to explain which tools, memory items, observations, or intermediate steps shaped an outcome and where failures originated \citep{wang2026agent}.
Outside LLM agents, chain-aware microservice tracing similarly shows that locally normal events can form structurally anomalous execution paths, a cross-domain analogy for local-validity versus global-failure diagnostics across execution traces and outputs \citep{xu2026chainawareencodingmicroservicetrace}.
These analyses make error type part of the evaluation object, not only a post-hoc anecdote.
For tool-using systems, that distinction is especially useful because a failed task can arise before semantic grading, at the native-call interface itself, or after a well-formed call has been produced.
Our taxonomy is narrower: it stays within native BFCL tool calls and separates structural malformed-output failures from well-formed but wrong tool-use failures.

\paragraph{Agent benchmarking and per-run consistency.}
Agent benchmarks add tool calls, intermediate traces, and environment interaction to ordinary input-output evaluation. Repository-level code-agent work gives adjacent examples: PerfAgent argues that passing tests is not enough for optimization agents and uses profiler- and verifier-guided feedback to refine beyond the first passing patch, while Lessons Learned reports category-specific code-optimization strengths and learning from successful and failed agent attempts \citep{deng2026perfagent,liu2025lessonslearned}. Tau-bench explicitly reports pass$^{k}$-style consistency across repeated trials, which makes per-task reliability visible rather than reducing every task to one run \citep{yao2024taubench}. That consistency view asks how often an agent succeeds at least once or repeatedly over multiple trials. Our audit instead estimates score-scale movement under frozen reruns and matched prompt perturbations. Function-calling leaderboards build on the tool-use evaluation setting introduced by Gorilla and operationalized in BFCL \citep{patil2024gorilla,bfcl2026leaderboard}.
Broader tool-use benchmarks such as ToolLLM/ToolBench evaluate API planning, retrieval, invocation, and generalization over large tool collections \citep{qin2023toolllm}.
Adjacent Text-to-SQL work similarly frames modern database interfaces as autonomous and interactive LLM pipelines involving iterative refinement and multi-agent collaboration \citep{su2026agentic}. That work studies task architectures and database-agent capabilities, whereas our audit studies measurement variability in a fixed native tool-calling benchmark harness.

\section{Experimental Setup}

\paragraph{Suite and study set.}
The audit uses the official Berkeley Function Calling Leaderboard (BFCL) distribution \citep{bfcl2026dataset}. We restrict the study to the AST-graded \texttt{multiple} and \texttt{parallel} categories. The frozen study set contains \result{study_matched_instances} matched instances split across \result{study_category_count} categories, and the sorted instance-id list is hash-frozen before any reported run. Any unrecoverable unit is handled by dropping the whole instance across all run indices so paired comparisons remain matched.
All reported endpoint-arm cells retain \result{study_matched_instances} matched instances; the exclusion table contains \result{study_excluded_instances} excluded instances and \result{study_unrecoverable_units} unrecoverable units.

\paragraph{Interface and scoring.}
All endpoints are called through their native tool or function-calling interface. Each BFCL instance supplies the official user request and official tool schema; an empty schema is a harness error before any provider call. The score is AST exactness against the official \texttt{ground\_truth}: the predicted function-name multiset must match, and each expected argument must match one of the allowed ground-truth values. Parse failures, malformed provider tool payloads, version-drift aborts, and truncations are recorded separately and count as incorrect unless the analysis explicitly excludes an unrecoverable instance under the paired-design policy.

\paragraph{Endpoints.}
The reported endpoints are \texttt{llama-3.1-8b-instant}, \texttt{llama-3.3-70b-versatile}, and \texttt{gemini-3.5-flash}. Gemini calls pin \texttt{thinkingLevel} to \texttt{low}; retained main-arm Gemini calls average \result{bfcl_gemini_mean_thoughts_tokens} thinking tokens. For every call, the raw row records the configured model identifier, the served model identity returned by the provider, and provider metadata. The version-drift guard aborts an experiment if the served model identity changes for a configured endpoint. Groq \texttt{system\_fingerprint} is recorded as a covariate rather than an abort trigger.

\paragraph{Arms.}
The main arm evaluates each endpoint on all \result{study_matched_instances} frozen instances with \(N=\result{study_main_reruns}\) reruns at temperature \result{study_main_temperature}. The perturbation arm uses the same \result{study_matched_instances} frozen instances, \(N=\result{study_perturbation_reruns}\) reruns, and \result{study_perturbation_count} named semantics-preserving prompt variants for each endpoint. The perturbations are designed to preserve the requested tool-call semantics while changing prompt surface form.

\paragraph{Perturbation definitions.}
Table~\ref{tab:perturbation-definitions} lists the perturbation variants.
Each variant changes only the user-message surface form; the BFCL tool schema, target function set, and ground-truth call are held fixed.

\begin{table*}[t]
  \centering
  \small
  \setlength{\tabcolsep}{4pt}
  \caption{Prompt perturbations used in the perturbation arm.}
  \label{tab:perturbation-definitions}
  \begin{tabular}{p{0.24\textwidth}p{0.37\textwidth}p{0.31\textwidth}}
    \toprule
    Variant & Surface change & Why semantics are preserved \\
    \midrule
    \texttt{compact\_user\_request} & Collapse user-message whitespace while preserving token order. & The requested operation and arguments are unchanged. \\
    \texttt{tool\_instruction\_prefix} & Prefix the user message with a generic instruction to use provided tools when needed. & The prefix does not add, remove, or alter any requested operation. \\
    \texttt{request\_label\_wrap} & Wrap the original request in neutral request-boundary labels. & The wrapped request text is unchanged. \\
    \texttt{call\_only\_suffix} & Append a generic instruction to answer through the function-calling interface rather than prose. & BFCL scoring already evaluates the requested tool call. \\
    \bottomrule
  \end{tabular}
\end{table*}

\paragraph{Reported uncertainty.}
We report both \marginalsd{} across runs and \pairedsd{} on matched instances. The marginal quantity summarizes the dispersion of aggregate scores across reruns. The paired quantity first computes matched score differences on the same instances and then reports the standard deviation of those paired differences. This is the relevant quantity for a reported difference \(A-B\): when aggregate measurements are nearly independent, the difference follows the standard independent-measurement approximation. The observed paired-to-\(\sqrt{2}\)-marginal ratios under that approximation are \result{bfcl_8b_paired_over_sqrt2_marginal_ratio}, \result{bfcl_70b_paired_over_sqrt2_marginal_ratio}, and \result{bfcl_gemini_paired_over_sqrt2_marginal_ratio}. These ratios are close to unity, indicating that reruns are nearly independent at each endpoint, so the paired quantity is approximately \(\sqrt{2}\) times the marginal SD.

\paragraph{Design isolation.}
The paired quantity is also the right target for any reported leaderboard gap \(A-B\), because the uncertainty of a difference is the uncertainty of matched score changes rather than the variability of either endpoint in isolation; the observed paired-to-\(\sqrt{2}\)-marginal ratios above verify that the usual independent-measurement approximation is adequate for these reruns.
Freezing the instance set removes instance-mix variance from the reported comparisons: each endpoint, rerun, and perturbation is evaluated on the same \result{study_matched_instances} instances, so the residual movement being estimated is rerun or prompt-perturbation movement on fixed cases.
This matters because BFCL categories are heterogeneous: changing which instances appear in a slice can move an aggregate score even when the endpoint behavior on every fixed instance is unchanged.
The design therefore separates finite-sample composition from repeated-measurement variability, and it makes the perturbation arm a within-instance robustness test rather than a comparison between different prompt subsets.
Finally, temperature \result{study_main_temperature} is a conservative measurement target for deployed tool use: any deployment that samples at temperature above \result{study_main_temperature} introduces an additional randomness source, so its total run-to-run variability should not be assumed to be smaller than the floor measured here.
The reported floor should therefore be read as the residual left after the obvious sampling knob has been set to its most deterministic value, not as a guarantee for ordinary production settings.

\paragraph{Data flow.}
Raw per-call outputs and explicit parse-failure rows are retained for auditability.

\section{Results}

\paragraph{Rerun near-determinism.}
At temperature \result{study_main_temperature}, all \result{study_endpoint_count} endpoints are close to deterministic on the frozen BFCL subset.
The mean pairwise run correlations are \result{bfcl_8b_mean_pairwise_run_correlation} for \texttt{llama-3.1-8b-instant}, \result{bfcl_70b_mean_pairwise_run_correlation} for \texttt{llama-3.3-70b-versatile}, and \result{bfcl_gemini_mean_pairwise_run_correlation} for \texttt{gemini-3.5-flash}.
The corresponding ever-flip fractions are \result{bfcl_8b_ever_flip_fraction}, \result{bfcl_70b_ever_flip_fraction}, and \result{bfcl_gemini_ever_flip_fraction}.
On the reported score scale, rerun \pairedsd{} is \result{bfcl_8b_paired_noise_floor_pp}, \result{bfcl_70b_paired_noise_floor_pp}, and \result{bfcl_gemini_paired_noise_floor_pp}.

\begin{table*}[t]
  \centering
  \scriptsize
  \setlength{\tabcolsep}{3pt}
  \caption{Rerun determinism on the frozen BFCL subset.}
  \label{tab:determinism}
\begin{tabular}{llllllll}
\toprule
Endpoint & Corr. & Always C & Always W & Flip inst. & Flip frac. & Marg. SD & Paired SD \\
\midrule
Gemini & 0.961286 & 118 & 28 & 4 & 0.0266667 & 0.00798146 & 0.0112954 \\
Groq 8B & 0.996744 & 107 & 42 & 1 & 0.00666667 & 0.00210819 & 0.00282684 \\
Groq 70B & 0.966475 & 121 & 26 & 3 & 0.02 & 0.00706233 & 0.00911825 \\
\bottomrule
\end{tabular}

\end{table*}

\paragraph{Perturbation is the larger floor.}
The perturbation arm changes prompt surface form while preserving the requested tool-call semantics.
Across \result{study_perturbation_count} variants, marginal accuracy is nearly constant: \result{bfcl_8b_perturbation_accuracy_range_pct} for \texttt{llama-3.1-8b-instant}, \result{bfcl_70b_perturbation_accuracy_range_pct} for \texttt{llama-3.3-70b-versatile}, and \result{bfcl_gemini_perturbation_accuracy_range_pct} for \texttt{gemini-3.5-flash}.
The paired perturbation SDs are much larger, spanning \result{bfcl_8b_perturbation_paired_sd_range_pp}, \result{bfcl_70b_perturbation_paired_sd_range_pp}, and \result{bfcl_gemini_perturbation_paired_sd_range_pp}.
This pattern means the perturbations mostly reshuffle which matched instances pass, rather than moving the endpoint mean by a large amount.
The median perturbation paired SD is \result{bfcl_8b_median_perturbation_over_rerun_paired_sd_prose} the rerun paired SD for \texttt{llama-3.1-8b-instant}, \result{bfcl_70b_median_perturbation_over_rerun_paired_sd_prose} for \texttt{llama-3.3-70b-versatile}, and \result{bfcl_gemini_median_perturbation_over_rerun_paired_sd_prose} for \texttt{gemini-3.5-flash}.

\begin{table*}[t]
  \centering
  \small
  \setlength{\tabcolsep}{3pt}
  \caption{Pairwise prompt-perturbation matrix on matched instances.}
  \label{tab:perturbation-matrix}
\begin{tabular}{lllllll}
\toprule
Endpoint & Left & Right & Left acc. & Right acc. & Matched & Paired SD \\
\midrule
Gemini & compact & prefix & 0.806667 & 0.814667 & 750 & 0.163212 \\
Gemini & compact & wrap & 0.806667 & 0.817333 & 750 & 0.206421 \\
Gemini & compact & suffix & 0.806667 & 0.797333 & 750 & 0.196548 \\
Gemini & prefix & wrap & 0.814667 & 0.817333 & 750 & 0.163387 \\
Gemini & prefix & suffix & 0.814667 & 0.797333 & 750 & 0.174375 \\
Gemini & wrap & suffix & 0.817333 & 0.797333 & 750 & 0.208945 \\
Groq 8B & compact & prefix & 0.714667 & 0.733333 & 750 & 0.144958 \\
Groq 8B & compact & wrap & 0.714667 & 0.726667 & 750 & 0.158817 \\
Groq 8B & compact & suffix & 0.714667 & 0.721333 & 750 & 0.0964432 \\
Groq 8B & prefix & wrap & 0.733333 & 0.726667 & 750 & 0.182574 \\
Groq 8B & prefix & suffix & 0.733333 & 0.721333 & 750 & 0.167013 \\
Groq 8B & wrap & suffix & 0.726667 & 0.721333 & 750 & 0.186238 \\
Groq 70B & compact & prefix & 0.817333 & 0.818667 & 750 & 0.131737 \\
Groq 70B & compact & wrap & 0.817333 & 0.818667 & 750 & 0.109609 \\
Groq 70B & compact & suffix & 0.817333 & 0.818667 & 750 & 0.109609 \\
Groq 70B & prefix & wrap & 0.818667 & 0.818667 & 750 & 0.0895024 \\
Groq 70B & prefix & suffix & 0.818667 & 0.818667 & 750 & 0.0730784 \\
Groq 70B & wrap & suffix & 0.818667 & 0.818667 & 750 & 0.0516742 \\
\bottomrule
\end{tabular}

\end{table*}

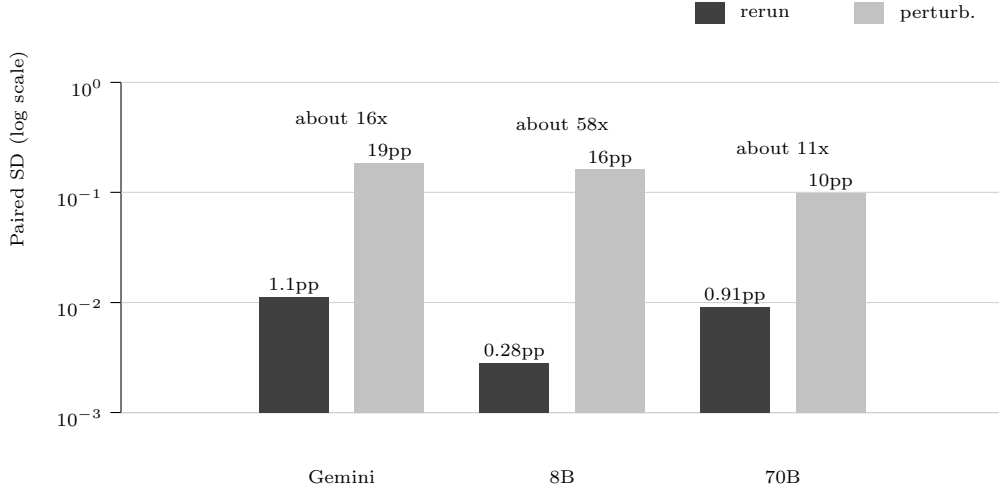
\begin{figure*}[t]
  \centering
\begingroup
\setlength{\unitlength}{1pt}
\begin{picture}(440,202)
\put(58.0,38.0){\line(1,0){332.0}}
\put(58.0,38.0){\line(0,1){124.0}}
\put(16.0,100.0){\rotatebox{90}{\scriptsize Paired SD (log scale)}}
\put(274.0,184.0){\textcolor{black!75}{\rule{11pt}{8pt}}}
\put(291.0,188.0){\makebox(0,0)[l]{\scriptsize rerun}}
\put(334.0,184.0){\textcolor{black!24}{\rule{11pt}{8pt}}}
\put(351.0,188.0){\makebox(0,0)[l]{\scriptsize perturb.}}
\put(58.0,38.0){\textcolor{black!18}{\line(1,0){332.0}}}
\put(54.0,38.0){\line(1,0){4}}
\put(51.0,36.0){\makebox(0,0)[r]{\scriptsize $10^{-3}$}}
\put(58.0,79.3){\textcolor{black!18}{\line(1,0){332.0}}}
\put(54.0,79.3){\line(1,0){4}}
\put(51.0,77.3){\makebox(0,0)[r]{\scriptsize $10^{-2}$}}
\put(58.0,120.7){\textcolor{black!18}{\line(1,0){332.0}}}
\put(54.0,120.7){\line(1,0){4}}
\put(51.0,118.7){\makebox(0,0)[r]{\scriptsize $10^{-1}$}}
\put(58.0,162.0){\textcolor{black!18}{\line(1,0){332.0}}}
\put(54.0,162.0){\line(1,0){4}}
\put(51.0,160.0){\makebox(0,0)[r]{\scriptsize $10^{0}$}}
\put(110.0,38.0){\textcolor{black!75}{\rule{26.0pt}{43.5pt}}}
\put(123.0,85.5){\makebox(0,0){\scriptsize 1.1pp}}
\put(146.0,38.0){\textcolor{black!24}{\rule{26.0pt}{93.8pt}}}
\put(159.0,135.8){\makebox(0,0){\scriptsize 19pp}}
\put(141.0,148.8){\makebox(0,0){\scriptsize \result{bfcl_gemini_median_perturbation_over_rerun_paired_sd_prose}}}
\put(141.0,14){\makebox(0,0){\scriptsize Gemini}}
\put(193.0,38.0){\textcolor{black!75}{\rule{26.0pt}{18.7pt}}}
\put(206.0,60.7){\makebox(0,0){\scriptsize 0.28pp}}
\put(229.0,38.0){\textcolor{black!24}{\rule{26.0pt}{91.4pt}}}
\put(242.0,133.4){\makebox(0,0){\scriptsize 16pp}}
\put(224.0,146.4){\makebox(0,0){\scriptsize \result{bfcl_8b_median_perturbation_over_rerun_paired_sd_prose}}}
\put(224.0,14){\makebox(0,0){\scriptsize 8B}}
\put(276.0,38.0){\textcolor{black!75}{\rule{26.0pt}{39.7pt}}}
\put(289.0,81.7){\makebox(0,0){\scriptsize 0.91pp}}
\put(312.0,38.0){\textcolor{black!24}{\rule{26.0pt}{82.6pt}}}
\put(325.0,124.6){\makebox(0,0){\scriptsize 10pp}}
\put(307.0,137.6){\makebox(0,0){\scriptsize \result{bfcl_70b_median_perturbation_over_rerun_paired_sd_prose}}}
\put(307.0,14){\makebox(0,0){\scriptsize 70B}}
\end{picture}
\endgroup
  \caption{Rerun paired SD versus median perturbation paired SD for each endpoint.}
  \label{fig:rerun-vs-perturbation}
\end{figure*}

\paragraph{Bootstrap sample-size curve.}
The bootstrap curve recomputes rerun paired SD over \result{sd_curve_n_instance_min} to \result{sd_curve_n_instance_max} matched instances with \result{sd_curve_bootstrap_replicates} resamples per endpoint.
The estimates remain small throughout the curve, with endpoint-specific means at the full matched size of \result{bfcl_8b_sd_curve_full_n_mean_pp}, \result{bfcl_70b_sd_curve_full_n_mean_pp}, and \result{bfcl_gemini_sd_curve_full_n_mean_pp}.
The full curve table is in Appendix~\ref{sec:appendix-diagnostics}; it shows why very small matched subsets should not be used to estimate this floor: the bootstrap intervals widen substantially at the low-instance end, so a rerun-SD estimate is more trustworthy only once enough matched instances are included to stabilize the resampling curve.
The practical implication is that a small pilot can reveal whether rerun noise exists, but it should not be treated as a precise endpoint-ranking instrument until the matched sample is large enough for the bootstrap intervals to settle.

\paragraph{Endpoint fingerprint variance.}
System fingerprints do not explain away the rerun floor in this audit.
When fingerprints vary, within-fingerprint and across-fingerprint paired SDs remain on the same order (reported at the per-pair instance level; see the Table A.2 caption); Gemini reports \result{bfcl_gemini_distinct_fingerprints} distinct fingerprint in these logs.
The full fingerprint table is in Appendix~\ref{sec:appendix-diagnostics}; it supports a negative conclusion: backend routing, as proxied by the recorded system fingerprint, is not the dominant explanation for the observed rerun floor in this harness.
The same-order within/across pattern is important because a routing explanation would predict most paired variation to appear across fingerprint changes, rather than persisting inside fingerprint-matched comparisons.

\paragraph{Capability does not predict stability here.}
The larger Groq endpoint is noisier under frozen reruns than the smaller Groq endpoint: \result{bfcl_70b_paired_noise_floor_pp} versus \result{bfcl_8b_paired_noise_floor_pp} paired SD.
Gemini is close to the larger Groq endpoint on rerun floor, with \result{bfcl_gemini_paired_noise_floor_pp} paired SD, while having a different provider and thinking-enabled configuration.
Under prompt perturbations, the direction across the Groq pair reverses: the median perturbation paired SD is \result{bfcl_70b_median_perturbation_paired_sd_pp} for \texttt{llama-3.3-70b-versatile} and \result{bfcl_8b_median_perturbation_paired_sd_pp} for \texttt{llama-3.1-8b-instant}; Gemini is \result{bfcl_gemini_median_perturbation_paired_sd_pp}.
Within these endpoints and this suite, endpoint capability therefore does not consistently predict either rerun stability or perturbation stability, and the two stability metrics rank the Groq pair in opposite orders.

\paragraph{Failure character shifts across endpoints.}
The taxonomy is defined at the native tool-call boundary.
\texttt{wrong\_function} means the predicted function-name multiset differs from the expected function-name multiset.
\texttt{wrong\_arguments} means the function names are right but required arguments are missing, mistyped, or have values outside the accepted ground-truth set.
\texttt{malformed\_output} means the provider output or tool-call payload is structurally unparseable, missing, truncated, or degenerate before ordinary argument grading.
\texttt{other} is the residual class for well-formed task failures not covered above, chiefly wrong call counts.
As an illustrative malformed-output case, a Gemini retry produced an over-long mixed-script token-soup argument, redacted here as \texttt{arg: "<mixed-script"[...]}; the provider rejected it before it could be graded as an ordinary wrong argument, and the retained taxonomy table records such cases as \texttt{truncated\_or\_abnormal\_finish}.
Malformed-output failures account for \result{bfcl_8b_malformed_output_failure_rate} of task failures on \texttt{llama-3.1-8b-instant}, \result{bfcl_70b_malformed_output_failure_rate} on \texttt{llama-3.3-70b-versatile}, and \result{bfcl_gemini_malformed_output_failure_rate} on \texttt{gemini-3.5-flash}.
The weaker endpoint therefore has a visibly larger structural-failure share, while the stronger Groq endpoint and the thinking-enabled Gemini endpoint fail mostly with well-formed but wrong arguments or wrong call counts.
Marginal accuracy hides this change in failure character.

\begin{table}[t]
  \centering
  \scriptsize
  \setlength{\tabcolsep}{3pt}
  \caption{Main-arm task-failure taxonomy on deduplicated retained rows. Counts are over the 150 x 10 main-arm reruns per endpoint; 'deduplicated' removes duplicate raw retry rows.}
  \label{tab:failure-taxonomy}
  \resizebox{\columnwidth}{!}{
\begin{tabular}{llllll}
\toprule
Endpoint & Wrong fn. & Wrong args & Malformed & Other & Malformed share \\
\midrule
Gemini & 0 & 263 & 2 & 36 & \textless{}1\% \\
Groq 8B & 10 & 270 & 130 & 19 & 30\% \\
Groq 70B & 0 & 247 & 20 & 10 & 7\% \\
\bottomrule
\end{tabular}
}
\end{table}

\section{Discussion}

\paragraph{Practical implication.}
For these endpoints on the frozen BFCL subset, extra reruns at the frozen decoding setting buy little precision relative to their cost.
The observed rerun floor is small: paired SDs are \result{bfcl_8b_paired_noise_floor_pp}, \result{bfcl_70b_paired_noise_floor_pp}, and \result{bfcl_gemini_paired_noise_floor_pp}, and ever-flip fractions remain \result{bfcl_8b_ever_flip_fraction}, \result{bfcl_70b_ever_flip_fraction}, and \result{bfcl_gemini_ever_flip_fraction}.
The uncertainty that remains visible in this audit is instead prompt-template robustness: median perturbation paired SDs are \result{bfcl_8b_median_perturbation_paired_sd_pp}, \result{bfcl_70b_median_perturbation_paired_sd_pp}, and \result{bfcl_gemini_median_perturbation_paired_sd_pp}.

\paragraph{Operational implication.}
At temperature \result{study_main_temperature} on native tool calling, repeated runs buy almost no extra information after the first small rerun check, so evaluation compute is better spent on matched prompt perturbations, grader audits, or broader instance coverage.
Because the observed perturbation multiples are \result{bfcl_8b_median_perturbation_over_rerun_paired_sd_prose}, \result{bfcl_70b_median_perturbation_over_rerun_paired_sd_prose}, and \result{bfcl_gemini_median_perturbation_over_rerun_paired_sd_prose} the rerun floor, leaderboard gaps should report prompt sensitivity alongside any rerun error bar.
The failure taxonomy adds a second operational warning: marginal accuracy alone hides structural fragility in weaker endpoints, where malformed-output failures remain a larger share of task failures than on the stronger Groq endpoint or the thinking-enabled Gemini setting.
For benchmark maintainers, this suggests a reporting contract: publish the frozen instance list, the prompt-template family, paired perturbation SDs, and a small failure-character table with each headline score.
For model users, the same contract turns a leaderboard gap into a deployment question: if a gap is smaller than the prompt-perturbation floor, prompt wording and call-surface robustness are more actionable than another batch of identical reruns.
For model developers, the failure taxonomy changes what should be debugged: malformed calls point to interface reliability and decoding constraints, whereas well-formed but wrong arguments point to semantic tool-use behavior.
Complementary agent-reliability work uses transactional belief commits, dependency-guided rollback repair, provenance-aware shared memory, or embodied recovery-augmented task graphs to diagnose or contain failures during longer-horizon execution \citep{li2026memtx,yu2026faulty,wang2026mapgraph,xu2026reaction}.
NFA instead operates at the benchmark measurement layer, using retained native outputs to separate rerun noise, prompt-surface sensitivity, and tool-call failure character.

\paragraph{Methodological implication.}
Marginal accuracy alone hides distinct layers of measurement behavior.
It hides rerun stability: the endpoint means can look like ordinary benchmark scores even when only a small set of instances ever changes outcome.
It hides the perturbation floor: per-variant accuracy ranges are narrow, but matched perturbation paired SDs are much larger than rerun paired SDs.
It also hides failure character: malformed-output shares move from \result{bfcl_8b_malformed_output_failure_rate} to \result{bfcl_70b_malformed_output_failure_rate} to \result{bfcl_gemini_malformed_output_failure_rate}, while many remaining task failures are well-formed calls with wrong arguments or wrong call counts.
For score differences reported on this suite, the paired perturbation matrix and failure taxonomy are therefore more informative than a single marginal accuracy.

\paragraph{Open question.}
The relationship between capability and stability is not consistent in this measurement.
\texttt{llama-3.3-70b-versatile} has the larger rerun paired SD than \texttt{llama-3.1-8b-instant}, but the smaller median perturbation paired SD.
Gemini adds a cross-provider, thinking-enabled point rather than a clean scale comparison.
This audit does not explain the reversal; it only establishes that, for these endpoints and this suite, apparent endpoint capability is not the same thing as uniformly lower measurement variability.

\section{Limitations}

\paragraph{Scope limits.}
This is a scope audit for a specific harness snapshot, not a stable public leaderboard.
The benchmark scope is \result{study_suite_count} suite: BFCL native tool calling on the AST-graded \texttt{multiple} and \texttt{parallel} categories, with no executable BFCL categories, irrelevance categories, retrieval tasks, or multi-turn agent trajectories.
The result should therefore not be generalized to agents whose errors emerge through environment state, tool side effects, retrieval quality, or long-horizon planning.
The completed results cover \result{study_endpoint_count} endpoints from \result{study_provider_count} providers; this supports a cross-provider scope statement, but it is still too small to estimate provider-level variance.
The endpoint identifiers are also time-bound API products rather than immutable model artifacts.
The version-drift guard protects this audit from within-study served-version changes, but it cannot make future provider deployments reproduce the same floor.
The reported main and perturbation arms primarily use temperature \result{study_main_temperature}; a temperature-\result{study_unreported_temperature_arm} arm exists outside the completed results reported here, so the paper should not be read as a temperature-sensitivity study.
Gemini differs from the Groq endpoints on provider and thinking configuration simultaneously, and its main-arm calls average \result{bfcl_gemini_mean_thoughts_tokens} thinking tokens, so this design cannot separate provider effects from the thinking-enabled setting.
The graders are programmatic AST matchers: they make the audit repeatable, but they do not execute calls and can miss semantically correct calls outside the accepted normalization rules or over-credit calls that match the expected structure for superficial reasons.
The failure taxonomy inherits that grader boundary: it classifies failures in the retained native-call outputs, not latent reasoning errors that would require human adjudication or executable task replay.
The taxonomy is also interface-specific.
A malformed native tool payload, a missing provider tool call, and a wrong but parseable argument are operationally different for an application, but this audit does not claim that the same categories would transfer unchanged to JSON-only prompting, constrained decoding, or agent frameworks with repair loops.
Finally, the perturbation set is non-archival: the \result{study_perturbation_count} named variants expose prompt-template fragility in this setup, but they do not span every reasonable prompt rewrite, system-message policy, tool-instruction convention, or application wrapper.
The perturbation results should therefore be read as a lower-level robustness audit over a declared prompt family, not as an estimate over the population of all possible prompts.

\paragraph{Statistical and causal limits.}
All estimates are conditional on the frozen study set and the endpoint behavior observed during this audit.
The matched design deliberately removes instance-mix variance from the main claims, but that also means the reported rerun and perturbation floors are not estimates of variability over every BFCL item, every future BFCL release, or every deployment prompt distribution.
The bootstrap curve in the appendix is a planning diagnostic for matched-set size; it is not a claim that the selected tasks are an independent sample from a stable population of tool-use problems.
The analysis is likewise descriptive rather than causal.
It can show that rerun noise is small, prompt perturbations are larger, and failure character differs across endpoints, but it cannot identify which training choices, routing policies, decoding constraints, or provider-side validators caused those patterns.
Finally, the audit treats the provider endpoint as the measurement object.
That is appropriate for leaderboard users who call hosted APIs, but it leaves out questions that require model weights, decoder internals, log probabilities, controlled seeds, or ablations of the serving stack.

\clearpage
\bibliography{refs}

\clearpage
\onecolumn
\appendix
\section{Diagnostic Tables}
\label{sec:appendix-diagnostics}
\setcounter{table}{0}
\renewcommand{\thetable}{A.\arabic{table}}

\begin{center}
  \setlength{\tabcolsep}{4pt}
  \renewcommand{\arraystretch}{1.12}
  \captionof{table}{Bootstrap rerun paired-SD curve by matched-instance count. These are bootstrap resampling means and can differ from the plug-in rerun paired-SD point estimates in Table 2 (0.28pp, 0.91pp, 1.1pp), because the bootstrap mean of a nonlinear statistic such as the SD need not equal the plug-in value.}
  \label{tab:sd-curve}
\begin{tabular}{llll}
\toprule
Endpoint & n & Mean SD & 95\% interval \\
\midrule
Gemini & 25 & 1.8pp & 0pp--6pp \\
Gemini & 50 & 1.6pp & 0pp--4.2pp \\
Gemini & 75 & 1.4pp & 0pp--3.4pp \\
Gemini & 100 & 1.4pp & 0pp--2.9pp \\
Gemini & 150 & 1.3pp & 0.4pp--2.6pp \\
Groq 8B & 25 & 0.33pp & 0pp--1.7pp \\
Groq 8B & 50 & 0.33pp & 0pp--1.7pp \\
Groq 8B & 75 & 0.36pp & 0pp--1.7pp \\
Groq 8B & 100 & 0.29pp & 0pp--0.85pp \\
Groq 8B & 150 & 0.26pp & 0pp--0.85pp \\
Groq 70B & 25 & 1.4pp & 0pp--5.9pp \\
Groq 70B & 50 & 1.2pp & 0pp--4.1pp \\
Groq 70B & 75 & 1.2pp & 0pp--3.2pp \\
Groq 70B & 100 & 1.1pp & 0pp--3pp \\
Groq 70B & 150 & 1.1pp & 0pp--2.4pp \\
\bottomrule
\end{tabular}

\end{center}

\begin{center}
  \setlength{\tabcolsep}{4pt}
  \renewcommand{\arraystretch}{1.12}
  \captionof{table}{Within- and across-fingerprint rerun paired differences. The SD columns report per-pair, instance-level paired-difference SDs across matched rerun pairs; these are at a different aggregation level from the aggregate-score paired SDs in Table 2 and are not directly comparable in magnitude.}
  \label{tab:fingerprint-variance}
\begin{tabular}{llllll}
\toprule
Endpoint & FPs & Within pairs & Within SD & Across pairs & Across SD \\
\midrule
Gemini & 1 & 6750 & 11pp & N/A & N/A \\
Groq 8B & 11 & 1887 & 5.1pp & 4863 & 2.9pp \\
Groq 70B & 8 & 1407 & 11pp & 5343 & 9.8pp \\
\bottomrule
\end{tabular}

\end{center}

\end{document}